\documentclass[runningheads]{llncs}
\usepackage[T1]{fontenc}
\usepackage{graphicx}
\usepackage{amsmath}
\usepackage{amsfonts}
\usepackage{csquotes}

\usepackage{caption}
\usepackage{subcaption}

\usepackage{bbding} 
\usepackage[misc]{ifsym}
\usepackage{xcolor}

\begin{document}
    \title{Evidence on the Regularisation Properties
    of Maximum-Entropy Reinforcement Learning}

    \titlerunning{Evidence on the Regularisation Properties
    of Maximum-Entropy RL}
%
%
    \author{Rémy Hosseinkhan Boucher\inst{1,2}\orcidID{0009-0005-3382-6081} \and \\
    Onofrio Semeraro\inst{1,2}\orcidID{0000-0002-0130-0545}
    \and
    Lionel Mathelin\inst{1,2}}
    \authorrunning{R. Hosseinkhan Boucher et al.}
%
    \institute{Université Paris-Saclay, Orsay, France \and CNRS,
        Laboratoire Interdisciplinaire des Sciences du Numérique,
        Orsay, France
        \email{\{remy.hosseinkhan,
            onofrio.semeraro,
            lionel.mathelin\}@upsaclay.fr}}
    \maketitle              
    \begin{abstract}
        The generalisation and robustness properties of policies learnt through
        Maximum-Entropy Reinforcement Learning are investigated
        on chaotic dynamical systems with Gaussian noise on the observable.
        First, the robustness under noise contamination of the agent's observation
        of entropy regularised policies is observed.
        Second, notions of statistical learning theory,
        such as complexity measures on the learnt model, are borrowed
        to explain and predict the phenomenon.
        Results show the existence of a relationship between entropy-regularised policy optimisation and robustness to noise, which can be described by the chosen complexity measures.

    \end{abstract}
%
%
%


    \section{Introduction}\label{sec:introduction}

    Maximum-Entropy Reinforcement Learning~\cite{williams1991} aims to solve the
    problem of learning a policy which optimises a chosen utility criterion
    while promoting the entropy of the policy.
    The standard way to account for the constraint is to add a Lagrangian term to the objective function.
    This entropy-augmented objective is commonly referred to as the soft objective.

    There are multiple advantages in solving the soft objective over
    the standard objective.
    For instance, favouring stochastic policies over deterministic ones
    allows learning multi-modal distributions~\cite{haarnoja2017}.
        {\color{black} In addition, agent stochasticity is a suitable way to deal
    with uncertainty induced by Partially
    Observable Markov Decision Processes (PO-MDP).}
        {\color{black}Indeed, there are PO-MDP such that the best
    stochastic adapted policy can be arbitrarily better than
    the best deterministic adapted policy~\cite{sigaud2010}\footnote{
        In this context, the term ``stochastic adapted policy'' is a conditional
        distribution on the control space $\mathcal{U}$ given
        the observation space $\mathcal{Y}$
        since this type of policy is ``adapted'' from Markovian policies
        in fully observable MDPs.}.} \\
    Furthermore, several important works highlight both theoretical
    and experimental \emph{robustness} of those policies under noisy dynamics
    and rewards~\cite{eysenbach2022}. 

    Related to the latter notion of robustness,
    the maximum-entropy principle exhibits non-trivial
    generalisation capabilities,
    which are desired in real-world applications~\cite{haarnoja2018}.

    However, the reasons for such robustness properties
    are not yet well understood.
    Thus, further investigations are needed to grasp the potential
    of the approach and to design endowed algorithms.
    A clear connection between Maximum-Entropy RL
    and their robustness properties is important and intriguing.

    Meanwhile, recent work in the deep learning community discusses
    how some complexity measures on the neural network model
    are related to generalisation, and explain typically
    observed phenomena~\cite{neyshabur2017}.
        {\color{black}In fact, these complexity measures are derived from the
    learnt model, bound the PAC-Bayes generalisation error, and are meant
    to identify which of the local minima generalise well.}

    As a matter of fact, a relatively recent trend in statistical learning
    suggests generalisation is not only favored
    by the regularisation techniques (\textit{e.g.}, dropout)
    but mainly because of the flatness
    of the local minima~\cite{hochreiter1997,dinh2017,keskar2017}.
    The reasons for such regularity properties remain an open problem.
    This work aims to address these points
    in the context of Reinforcement Learning,
    and addresses the following questions:
%
%

    \emph {What is the bias introduced by entropy regularisation?}
    \emph {Are the aforementioned complexity measures also related
    to the robustness of the learnt solutions
    in the context of Reinforcement Learning?}

    In that respect, by defining a notion of robustness
    against noisy contamination of the observable,
    a study on the impact of the entropy regularisation on the robustness
    of the learnt policies is first conducted.
    After explaining the rationale behind the choice of the complexity measures,
    a numerical study is performed to validate the hypothesis
    that some measures of complexity are good robustness predictors.
    Finally, a link between the entropy regularisation
    and the flatness of the local minima is treated
    through the information geometry notion of Fisher Information.

        {\color{black}
    The paper is organised as follows.
    Section~\ref{sec:related-work} introduces the background
    and related work, Section~\ref{sec:problem-setup-and-background}
    presents the problem setting.
    Section~\ref{sec:complexity-measures-and-robustness} is the core contribution of this paper.
    This section introduces
    the rationale behind the studied complexity measures from a learning theory perspective, as well as their expected relation to robustness. 
    Lastly, Section~\ref{sec:experiments} presents the experiments related
    to the policy robustness as well as their complexity,
    while Section~\ref{sec:results} examines the results obtained.
    Finally, Section~\ref{sec:discussion} concludes the paper.
    }

    \section{Related work}
    \label{sec:related-work}
    \paragraph[Maximum Entropy Policy Optimisation]
    {Maximum Entropy Policy Optimisation}
    \label{par:maximum-entropy-reinforcement-learning-related-work}

    {\color{black} In~\cite{haarnoja2018},
        the generalisation capabilities of entropy-based policies are observed
        where multimodal policies lead to optimal solutions.
        It is suggested that maximum entropy solutions
        aim to learn all the possible ways to solve a task.
        Hence, transfer learning to more challenging objectives
    is made easier, as demonstrated in their experiment.
    This study investigates the impact of adopting policies
    with greater randomness on their robustness.}
    The impact of the entropy regularisation on the loss landscape
    has been recently studied in~\cite{ahmed2019}.
    They provide experimental evidence about the smoothing effect of entropy
    on the optimisation landscape. 
        {\color{black} The present study aims specifically to answer the question
    in Section 3.2.4 of their paper:
    \emph{Why do high entropy policies learn better final solutions?}}
    This paper extends their results from a complexity measure point of view.
    Recently,~
    \cite{neu2017,derman2021} studied the equivalence
    between robustness and entropy regularisation on regularised MDP\@.

        {\color{black}
    \paragraph[Flat minima and Regularity]
    {Flat minima and Regularity}
    \label{par:flat-minima-and-regularity}
    The notion of local minima flatness was first introduced in the context of
    supervised learning by~\cite{hochreiter1997}
    through the Gibbs formalism~\cite{haussler}. 
    Progressively, different authors stated the concept
    with geometric tools such as first order (gradient)
    or second order (Hessian) regularity
    measures~\cite{zhao2022,keskar2017,sagun2017,yoshida2017,dinh2017}.
    In a similar fashion,~
    \cite{chaudhari2019} uses the concept of local entropy
    to smooth the objective function. \\
    In the scope of Reinforcement Learning,~
    \cite{ahmed2019} observed that flat minima
    characterise maximum entropy solutions,
    and entropy regularisation
    has a smoothing effect on the loss landscape,
    reducing the number of local optima.
    A central objective of this present study is to investigate
    this latter property further
    and relate it to the field of research on robust optimisation.
    Lastly, among the few recent studies on the learning
    and optimisation aspects of RL,~
    \cite{gogianu2021} shows how a well-chosen regularisation
    can be very effective for deep RL\@.
    Indeed, they explain that constraining the Lipschitz constant
    of only one neural network layer is enough
    to compete with state-of-the-art performances on a standard benchmark.}

        {\color{black}
    \paragraph[Robust Reinforcement Learning]
    {Robust Reinforcement Learning}
    \label{par:robust-reinforcement-learning}
    A branch of research related to this work is the study of robustness
    with respect to the uncertainty of the dynamics,
    namely \emph{Robust Reinforcement Learning} (Robust RL),
    which dates back to the 1970s~\cite{satia1973}.
    Correspondingly, in the field of control theory,
    echoes the notion of robust control
    and especially $H_\infty$ control~\cite{zhou1996},
    which also appeared in the mid-1970s
    after observing Linear Quadratic Regulator (LQR) solutions
    are very sensitive to perturbations
    while not giving consistent enough guarantees~\cite{doyle1996}. \\
    More specifically,
    the Robust RL paradigm aims to control the dynamics
    in the worst-case scenario, \textit{i.e.},
    to optimise the minimal performance
    for a given objective function
    over a set of possible dynamics through a min-max problem formulation.
    This set is often called \textit{ambiguity set} in the literature.
    It is defined as a region in the space of dynamics
    close enough w.r.t.\ to some divergence measure,
    such as the relative entropy~\cite{nilim2003}.
    Closer to this work, the recent paper from~\cite{eysenbach2022}
    shows theoretically how Maximum-Entropy RL policies are inherently robust
    to a certain class of dynamics of fully-observed MDP.\@
    The finding of their article might still hold
    in the partially observable setting
    as any PO-MDP can be cast as fully-observed MDP
    with a larger state-space of probability measures~\cite{hernandez-lerma1996},
    providing the ambiguity set is adapted to a more complicated space.}

    \section[Problem Setup and Background]
    {Problem Setup and Background}
    \label{sec:problem-setup-and-background}
    \subsection[Noisy Observable Markov Decision Process with Gaussian noise]
    {Partially Observable Markov Decision Process with Gaussian noise}
    \label{subsec:markov-decision-process}

    {\color{black} First, the control problem when noisy observations
    are available to the agent is formulated.}
    The study focuses on
    \emph{Partially Observable Markov Decision Processes (PO-MDP)}
    with Gaussian noise of the form~\cite{deisenroth2012}:
    \begin{equation}
        \begin{aligned}
            & X_{h+1}=F\left(X_{h},\, U_{h}\right) \\
            & Y_{h}=G\left(X_{h}\right)+ \epsilon,
            \quad \epsilon \sim \mathcal{N}(0, \sigma_Y^2 I_d) \\
        \end{aligned}
        \label{eq:partially-observable-markov-decision-process}
    \end{equation}
    with $X_{h} \in \mathcal{X}$, $U_{h} \in \mathcal{U}$
    and $Y_{h} \in \mathcal{Y}$ for any $h \in \mathbb{N}$,
    where $\mathcal{X}$, $\mathcal{U}$ and $\mathcal{Y}$
    are respectively the corresponding state, action and observation spaces.
    The initial state starts from a reference state $x_e^{*}$
    on which centred Gaussian noise with diagonal covariance $\sigma_{e}^2 I_d$
    is additively applied, $X_0 \sim \mathcal{N}(x_e^{*},\, \sigma_{e}^2 I_d)$.
    Associated with the dynamics,
    an instantaneous cost function
    $c: \mathcal{X} \times \mathcal{U} \rightarrow \mathbb{R}_+$
    is also given to define the control model.

    In this context, a \emph{policy} $\pi$
    is a transition kernel on $\mathcal{U}$ given $\mathcal{Y}$, \textit{i.e.},
    a distribution on actions conditioned on observations.
    This kind of policies are commonly used in the literature
    but can be very poor in the partially observable setting
    where information is missing.  
    Together, a control model, a policy $\pi$
    and an initial distribution $P_{X_0}$ on $\mathcal{X}$
    define a stochastic process with distribution $P^{\pi, \epsilon}$
    where the superscript $\epsilon$
    highlights the dependency on the observation noise $\epsilon$.
    Similarly, one denotes by $P^{\pi}$ the distribution of the process
    when the noise is zero almost-surely, \emph{i.e.}, $P^{\pi} = P^{\pi, 0}$.
        {\color{black} More details about the PO-MDP control problem
    can be found in~\cite{hernandez-lerma1996,cassandra1998}.}

    Here, the maximum-entropy control problem is to find a policy $\pi^*$
    which minimises the following performance criterion
    \begin{equation}
        J^{\pi, \epsilon}_m
        = \mathbb{E}^{\pi, \epsilon}
        \left[ \sum_{h=0}^{H} \gamma^{h} c\left(X_{h}, U_{h}\right) \right]
        + \alpha_m \mathbb{E}^{\pi, \epsilon}
        \left[ \sum_{h=0}^{H} \gamma^{h}
        \mathcal{H}(\pi(\, \cdot \mid X_h)) \right]
        \label{eq:finite-time-horizon-maximum-entropy-objective},
    \end{equation}
    where $H \in \mathbb{N}$ is a given time horizon,
    $\mathbb{E}^{\pi, \epsilon}$
    denotes the expectation under the probability measure $P^{\pi, \epsilon}$,
    $\mathcal{H}$ denotes the differential entropy~\cite{cover2006}
    and $\alpha_m$ is a time-dependent weighting parameter
    that evolves over training time $m \leq m_\mathcal{D} = |\mathcal{D}|$
    with $|\mathcal{D}|$ being the total number of times
    the agent interacts with the system
    such that all observations used by the learning algorithm
    form the dataset $\mathcal{D}$
    at the end of the training procedure
    (when $m_\mathcal{D}$ environment interactions are done). \\
    In the $\alpha_m = 0$ case,
    $J^{\pi, \epsilon}_m$ is denoted $J^{\pi, \epsilon}$.
    The quantity $J^{\pi, \epsilon}$ is called the value function or,
    more generally, \emph{loss}. \\
    Moreover,
    the performance gap for dynamics with noisy
    and noiseless observables will be considered
    in the sequel.
    In this context, the \emph{(rate of) excess risk under noise}
    is defined as the difference between the loss under noisy dynamics
    and the loss under noiseless dynamics:
    \begin{definition}[Excess Risk Under Noise]
        \label{def:excess-risk-under-noise}
        The excess risk under noise of a policy $\pi$
        for a PO-MDP
        with dynamics~\eqref{eq:partially-observable-markov-decision-process}
        is defined as:
        \begin{equation}
            \mathcal{R}^{\pi}
            = \mathbb{E}^{\pi, \epsilon}
            \left[ \sum_{h=0}^{H} \gamma^{h} c\left(X_{h}, U_{h}\right) \right]
            - \mathbb{E}^{\pi}
            \left[ \sum_{h=0}^{H} \gamma^{h} c\left(X_{h}, U_{h}\right) \right]
            = J^{\pi, \epsilon} - J^{\pi}
            \label{eq:excess-risk-under-noise}
        \end{equation}
        Similarly, the rate of excess risk under noise is defined as:
        \begin{equation}
            \mathring{\mathcal{R}}^{\pi}
            = \frac{J^{\pi, \epsilon} - J^{\pi}}{J^{\pi}}
            = \frac{\mathcal{R}^{\pi}}{J^{\pi}}
            \label{eq:rate-excess-risk-under-noise}
        \end{equation}
    \end{definition}
    Note that in the above definition,
    expectations are taken with respect to the probability measure
    $P^{\pi, \epsilon}$ and $P^{\pi}$ respectively.
    The \emph{rate of excess risk under noise}
    represents the performance degradation
    after noise introduction in value function units.
    In the rest of the paper,
    arguments to derive 
    complexity measures will be developed,
    allowing to predict the excess risk under noise
    and provide numerical evidence showing maximum-entropy policies
    are more robust regarding this metric.
        {\color{black} Hence, maximum-entropy policies
    implicitly learn a robust control policy
    in the sense of Definition~\ref{def:excess-risk-under-noise}.}

    In the next section,
    some concepts of statistical learning theory are introduced.
    Then, complexity measures will be defined to quantify the regularisation
    power of the maximum-entropy objective
    of~\eqref{eq:finite-time-horizon-maximum-entropy-objective}.

    \section[Complexity Measures and Robustness]
    {Complexity Measures and Robustness}
    \label{sec:complexity-measures-and-robustness}
    \subsection[Complexity measures]
    {Complexity Measures}
    \label{subsec:complexity-measures}

    The principal objective of \emph{statistical learning}
    is to provide bounds on the generalisation error,
    so-called \emph{generalisation bounds}. 
    In the following,
    it is assumed that an algorithm $\mathcal{A}$
    returns a hypothesis $\pi \in \mathcal{F}$ from a dataset $\mathcal{D}$.
        {\color{black} Note that the dataset $\mathcal{D}$ is random
    and the algorithm $\mathcal{A}$ is a randomised algorithm.}

    As the hypothesis set $\mathcal{F}$
    typically used in machine learning is infinite,
    a practical way to quantify the generalisation ability
    of such a set must be found.
    This quantification is done by introducing \emph{complexity measures},
    enabling the derivation of generalisation bounds.
    \begin{definition}[Complexity measure]
        A complexity measure is a mapping
        $\mathcal{M}: \mathcal{F} \rightarrow \mathbb{R}_+$
        that maps a hypothesis to a positive real number.
    \end{definition}
    According to~\cite{neyshabur2017} from which this formalism is inspired,
    an appropriate complexity measure satisfies several properties.
    In the case of parametric models $\pi_\theta \in \mathcal{F}(\Theta)$
    with $\theta \in \Theta \subset \mathbb{R}^b$,
    it should increase with the dimension $b$ of the parameter space $\Theta$
    as well as being able to identify
    when the dataset $\mathcal{D}$ contains totally random,
    spurious or adversarial data.
    As a result, finding good complexity measures $\mathcal{M}$
    allows the quantification of the generalisation ability of
    a hypothesis set $\mathcal{F}$ or a model $\pi$
    and an algorithm $\mathcal{A}$.

    \subsection[Complexity measures for PO-MDP with Gaussian Noise]
    {Complexity measures for PO-MDP with Gaussian Noise}
    \label{subsec:complexity-measures-for-po-mdp-with-gaussian-noise}


    This paper studies heuristics about generalisation bounds
    on the optimal \emph{excess risk under noise}
    from Definition~\ref{def:excess-risk-under-noise}
    when the optimal policy $\pi_{\theta^*}$ is learnt
    with an algorithm $\mathcal{A}$ on the non-noisy objective $J^{\pi}$,
    where $\alpha_m = 0$ for any $m$.
    \begin{definition}[(Rate of) Excess Risk Under Noise Bound]
        Given an optimal policy $\pi^*$ learnt with an algorithm $\mathcal{A}$
        on the non-noisy objective $J^{\pi}$,
        the optimal excess risk under noise bound
        is a real-valued mapping $\varphi$ such that
        \begin{equation}
            \mathcal{R}^{\pi^*}
            \leq
            \varphi(\mathcal{M}(\pi^*, \mathcal{D}),\,
            m_\mathcal{D},\, \eta,\, \delta)
            \label{eq:optimal-excess-risk-under-noise-bound}
        \end{equation}
        and $\varphi$ is increasing with the complexity measure $\mathcal{M}$
        and the sample complexity $m_\mathcal{D}$.
        The definition is similar for the rate of excess risk under noise bound
        where $\mathring{\mathcal{R}}^{\pi^*}$
        is used instead of $\mathcal{R}^{\pi^*}$.
    \end{definition}

    Hence, considering a learning algorithm $\mathcal{A}$
    with a parameterised family
    $\mathcal{F}(\Theta) = (\pi_\theta)_{\theta \in \Theta}$,
    $\Theta \subset \mathbb{R}^b$,
    such that $\theta = (\theta_\mu,\, \theta_{\sigma_\pi})$
    with $\pi_\theta(\cdot \mid x) \sim \mathcal{N}(\mu_{\theta_\mu}(x),\,
    \text{diag}(\theta_{\sigma_\pi}))$,
    $ x \in \mathcal{X}$,\
    - where $\mu_{\theta_{\mu}}$ is a shallow multi-layer
    feed-forward neural network
    (with depth-size $l = 2$,
    width $w = 64$ neurons,
    weights matrix $( \theta_\mu^i )_{1 \leq i \leq l}$)
    and $\text{diag}(\theta_{\sigma_\pi})$
    is a diagonal matrix of dimension $q = \dim(\mathcal{U})$ parameterising
    the variance\footnote{Note this choice of state-independent policy variance
    is inspired by~\cite{ahmed2019} and simplifies the problem.}
    - to learn the optimal policy $\pi_{\theta^*}$,
    multiple complexity measures $\mathcal{M}$
    are defined and details on their underlying rationale are given below.

    \subsubsection{Norm based complexity measures}
    \label{subsubsec:norm-based-complexity-measures}
    First, the so-called norm-based complexity measures are functions
    of the norm of some subset of the parameters of the model.
    For instance, a common norm-based measure calculates the product
    of the operator norms of the neural network linear layers.
    The measures are commonly used in the statistical learning
    theory literature to derive bounds on the generalisation gap,
    especially in the context
    of neural networks~\cite{neyshabur2015,golowich2018,miyato2018}. \\
    In fact, the product of the norm of the linear layers
    of a standard class of multi-layer neural networks
    (including Convolutional Neural Networks)
    serves as an upper bound on the often intractable
    Lipschitz constant of the network~\cite{miyato2018}.
    Thus, controlling the magnitude of the weights of the linear layers
    increases the regularity of the model. \\
    Consequently, the following complexity measures are defined:
    \begin{itemize}
        \item $\mathcal{M}(\pi_\theta, \mathcal{D}) = \| \theta_\mu \|_p$
    \end{itemize}
    \begin{itemize}
        \item $\mathcal{M}(\pi_\theta, \mathcal{D})
        = \Pi^l_{i=1} \| \theta_\mu^i \|_p$
        where $\theta_\mu^i$ is the $i^{th}$ layer
        of the network $\mu_{\theta_\mu}$.
    \end{itemize}
    In this context $\| \cdot \|_p$ with $p = 1,\, 2,\, \infty$
    denotes the $p$-operator norm while $p = F$ denotes the Frobenius norm,
    which is discarded for the first case
    of the full parameters vector $\theta_\mu$
    (since Frobenius norm is defined for matrix).

    \subsubsection{Flatness based complexity measures}
    \label{subsubsec:flatness-based-complexity-measures}
    On the other hand, another measure of complexity
    is given by the flatness of the optimisation local minimum
    (see Section~\ref{sec:related-work} for a brief overview).
    As~\cite{mcallester2003,neyshabur2017} have pointed out,
    the generalisation ability of a parametric solution
    is controlled by two key components in the context of supervised learning:
    the norm of the parameter vector
    and its flatness w.r.t.\ to the objective function.

    One might wonder if a similar robustness property
    still holds in the setting of Reinforcement Learning.
    In this manner,
    complexity measures quantifying the flatness of the solution are needed.
    Concretely,
    the interest lies in the flatness of the local minima
    of the objective function $J^{\pi}$.
    As stated earlier,
    there are several ways to quantify the flatness of a solution
    with metrics derived from the gradient or curvature
    of the loss function at the local optimum,
    such as the Hessian's largest eigenvalue
    - otherwise spectral norm~\cite{keskar2017}
    or the trace of Hessian~\cite{dinh2017}.

    Moreover, as discussed in Section~\ref{sec:related-work},~
    \cite{ahmed2019} observed that
    \emph{maximum entropy solutions are characterised by flat minima}
    while entropy regularisation has a smoothing effect on the loss landscape.
    \emph{Hence, a central objective of this present study is to
    investigate this latter property further and relate it
    to the robustness aspect of the resulting policies}.

    However,
    instead of dealing directly with the Hessian of the objective $J^\pi$
    this work proposes a measure based on
    the conditional Fisher Information $\mathcal{I}$ of the policy
    due to its link with a notion of model regularity
    in the parameter space.
    \begin{definition}[Conditional Fisher Information Matrix]
        Let $x \in \mathcal{X}$ and $\pi_\theta$ a policy identified
        by its conditional density
        for a parameter $\theta \in \Theta \subset \mathbb{R}^b$
        and suppose $\rho$ is a distribution over $\mathcal{X}$.
        The conditional Fisher Information Matrix of the vector $\theta$
        is defined \textit{under some regularity conditions} as
        \begin{equation}
            \mathcal{I}(\theta)
            = -\ \mathbb{E}^{X \sim \rho, U \sim \pi_\theta(\cdot | X)}
            \left[ \nabla^2_\theta \log \pi_\theta(U \mid X) \right]
            \label{eq:conditional-fisher-information},
        \end{equation}
        where $\nabla^2_\theta$ denotes the Hessian matrix
        evaluated at $\theta$.
    \end{definition}

    Note that the distribution over states $\rho$ is arbitrary and can be chosen
    as the discounted state visitation measure $\rho^\pi$
    induced by the policy $\pi$~\cite{agarwal2019}
    or the stationary distribution
    of the induced Markov process if the policy is Markovian
    and the MDP ergodic\footnote{With these choices,
        the following holds:
        $\mathbb{E}^{\rho^\pi(ds) \pi(da | s)} = \mathbb{E}^\pi$
        up to taking the expectation \textit{w.r.t.} the state-action space
        (no subscript under $X$ and $U$) or the trajectory space
        (with subscripts such as $X_h$ and $U_h$ as trajectory coordinate)
        \cite{agarwal2019}.}
    as it is done in~\cite{kakade2001}.

    As a matter of fact,
    it has already been mentioned
    in the early works of policy optimisation~\cite{kakade2001}
    that this quantity $\mathcal{I}$
    might be related to the Hessian of the objective function.
    Indeed, the Hessian matrix of the standard objective function reads
    (see~\cite{shen2019} for a proof):
    \begin{equation}
        \nabla_\theta^2{J^{\pi_\theta}}
        = \mathbb{E}^{\pi_\theta}
        \left[\sum_{h,i,j=0}^{H}
        c\left(X_h, U_h\right)
        \left( \Pi_1^{i,j}
        + \Pi_2^{i}
        \right) \right].
        \label{eq:finite-time-horizon-policy-hessian}
    \end{equation}
    where the second order quantities are given by
    \begin{align}
        \Pi_1^{i,j} &= \nabla_\theta \log \pi_\theta\left(U_i \mid X_i\right) \nabla_\theta \log \pi_\theta\left(U_j \mid X_j\right)^T \\
        \Pi_2^{i} &= \nabla_\theta^2 \left[ \log \pi_\theta\left(U_i \mid X_i\right)  \right]
        \label{eq:finite-time-horizon-policy-hessian_rhs}
    \end{align}
    As suggested by the author mentioned above
    (S. Kakade),~\eqref{eq:finite-time-horizon-policy-hessian}
    might be related to $\mathcal{I}$ although being weighted by the cost $c$.
    Indeed, the Hessian of the state-conditional log-likelihoods
    ($\nabla_\theta^2 \log \pi_\theta$ on the rightmost part of the expectation
    of~\eqref{eq:finite-time-horizon-policy-hessian})
    belongs to the objective-function Hessian $\nabla_\theta^2{J^{\pi_\theta}}$
    while the Fisher Information $\mathcal{I}(\theta)$
    is an average of the Hessian of the policy log-likelihood.

    In any case, the conditional FIM measures the regularity
    of a critical component of the objective to be minimised.
    Thus, the trace of the conditional FIM of the mean actor network parameter
    $\theta_\mu$ is suggested as a complexity measure


    \begin{itemize}
        \item $\mathcal{M}(\pi_\theta, \mathcal{D})
        = Tr( \mathcal{I}\left(\theta_\mu \right))
        = Tr( -\ \mathbb{E}^{X \sim \rho^\pi, U \sim \pi_\theta(\cdot | X)}
        \left[ \nabla^2_{\theta_\mu} \log \pi_\theta(U \mid X) \right])$.
    \end{itemize}
    Moreover, in the context of classification,
    a link between the degree of stochasticity of optimisation gradients
    (leading to flatter minima~\cite{mulayoff2020,xie2021})
    and the FIM trace during training
    has recently been revealed in~\cite{jastrzebski2021}.
        {\color{black} Magnitudes of the FIM eigenvalues
    may be related to loss flatness
    and norm-based capacity measures
    to generalisation ability~\cite{karakida2019} in deep learning.}

    \section[Experiments]{Experiments}\label{sec:experiments}

    \subsection[Robustness under noise of Maximum Entropy Policies]
    {Robustness under noise of Maximum Entropy Policies}
    \label{subsec:robustness-under-noise-of-maximum-entropy-policies}
    The first hypothesis is that maximum entropy policies
    are more robust to noise than those trained without entropy regularisation
    (which play the role of control experiments).
    Consequently, the robustness of the controlled policy $\pi_{\theta^*}$
    is compared with the robustness of the maximum entropy policy
    $\pi_{\theta^*}^{\alpha}$
    for different temperature evolutions
    $\alpha = (\alpha_m)_{0 \leq m \leq m_\mathcal{D}}$.
    In this view, and since inter-algorithm comparisons
    are characterised by
    high uncertainty~\cite{henderson2018,colas2018,agarwal2021},
    only one algorithm $\mathcal{A}$
    (\emph{Proximal Policy Optimisation} (PPO)~\cite{schulman2017})
    is retained while results on multiple entropy constraint levels
    $\alpha = (\alpha_m)_{0 \leq m \leq m_\mathcal{D}}$ are examined.

    In this regard, ten independent PPO models are trained for each of the five
    arbitrarily chosen entropy temperatures
    $\alpha^i = (\alpha^i_m)_{0 \leq m \leq m_\mathcal{D}}$
    where $i \in \{ 1, \dots , 5 \} $,
    on dynamics without observation noise,
    \textit{i.e.}, where $\sigma_Y^2 = 0$.
    The entropy coefficients linearly decay during training,
    and all vanish ($\alpha_m = 0$)
    when $m$ reaches one-fourth of the training time
    $m_{1/4} = \lfloor \frac{m_\mathcal{D}}{4} \rfloor$ in order to replicate
    a sort of exploration-exploitation procedure,
    ensuring that all objectives $J^\pi_m$
    are the same whenever $m \geq m_{1/4}$, \textit{i.e.}, $J^\pi_m = J^\pi$.
    This choice is different but inspired by~\cite{ahmed2019}
    as they optimise using only the \emph{policy gradient}
    and manipulate the standard deviation of Gaussian policies directly,
    whereas, in the present approach, it is done implicitly
    with an adaptive entropy coefficient.
    An algorithm that learns a model with a given entropy coefficient
    $\alpha = (\alpha_m)_{0 \leq m \leq m_\mathcal{D}}$
    is denoted as $\mathcal{A}_\alpha$.

    The chosen chaotic systems are the \emph{Lorenz}~\cite{vincent1991}
    (with $m_\mathcal{D} = 10^6$)
    and \emph{Kuramoto-Sivashinsky (KS)}~\cite{bucci2019}
    (with $m_\mathcal{D} = 2 \cdot 10^6$)
    controlled differential equations.
    The defaults training hyper-parameters
    from \textit{Stable-Baselines3}~\cite{raffin2021} are used.

    \subsection[Robustness against Complexity Measures]
    {Robustness against Complexity Measures}
    \label{subsec:robustness-against-complexity-measures}

    So far, three separate analyses on the 5 $\times$ 10 models obtained
    have been performed on the \emph{Lorenz}
    and \emph{Kuramoto-Sivashinsky (KS)} controlled differential equations. \\
    First, as mentioned before,
    the robustness of the models
    for each of the chosen entropy temperatures $\alpha^i$
    is tested against the same dynamics
    but now with a noisy observable, \textit{i.e.}, $\sigma_Y > 0$.
    Second, norm-based complexity measures introduced
    in Section~\ref{subsec:complexity-measures-for-po-mdp-with-gaussian-noise}
    are evaluated and compared to the generalisation performances
    of the distinct algorithms $\mathcal{A}_\alpha$.
    Third, numerical computation of the conditional distribution
    of the trace of the Fisher Information Matrix
    given by~\eqref{eq:conditional-fisher-information}
    is performed to test the hypothesis that this regularity measure
    is an indicator of robust solutions.
    The state distribution $\rho^{\pi_\theta}$
    is naturally chosen as the state visitation distribution
    induced by the policy $\pi_\theta$.
        {\color{black} The following section discusses
    the results of those experiments.}

    \section{Results}\label{sec:results}
    This section provides numerical evidence
    of maximum entropy's effect on the robustness,
    as defined by the Excess Risk Under Noise
    defined by~\eqref{eq:excess-risk-under-noise}.
    Then, after quantifying robustness,
    the relation between the complexity measures defined
    in Section~\ref{subsec:complexity-measures-for-po-mdp-with-gaussian-noise}
    and robustness is studied.

    \subsection[Entropy Regularisation induces noise robustness]
    {Entropy Regularisation induces noise robustness}
    \label{subsec:entropy-regularisation-induces-noise-robustness}
    In the first place,
    a distributional representation\footnote{By replacing
    the expectation operator $\mathbb{E}$
        with the conditional expectation $\mathbb{E}[~\cdot \mid X_0]$
        in the definition of $\mathcal{R}^\pi$
        in \eqref{eq:excess-risk-under-noise},
        the quantity becomes a random variable for which the distribution
        can be estimated by sampling
        the initial state distribution $X_0 \sim \mathcal{N}(x_e^{*},\,
        \sigma_{e}^2 I_d)$.
        In fact, taking the conditional expectation gives the difference
        of the standard \emph{value functions}
        under $P^\pi$ and $P^{\pi, \epsilon}$.}
    of the rate of excess risk under noise
    defined in~\eqref{eq:excess-risk-under-noise} is computed
    for each of the 5 $\times$ 10 models obtained with the PPO algorithm
    $\mathcal{A}_{\alpha^i}$, $i \in \{1,\ldots,5\}$
    and different levels of observation noise $\sigma_Y > 0$.

\begin{figure}[t]
        \centering
        \includegraphics[scale=0.275]
        {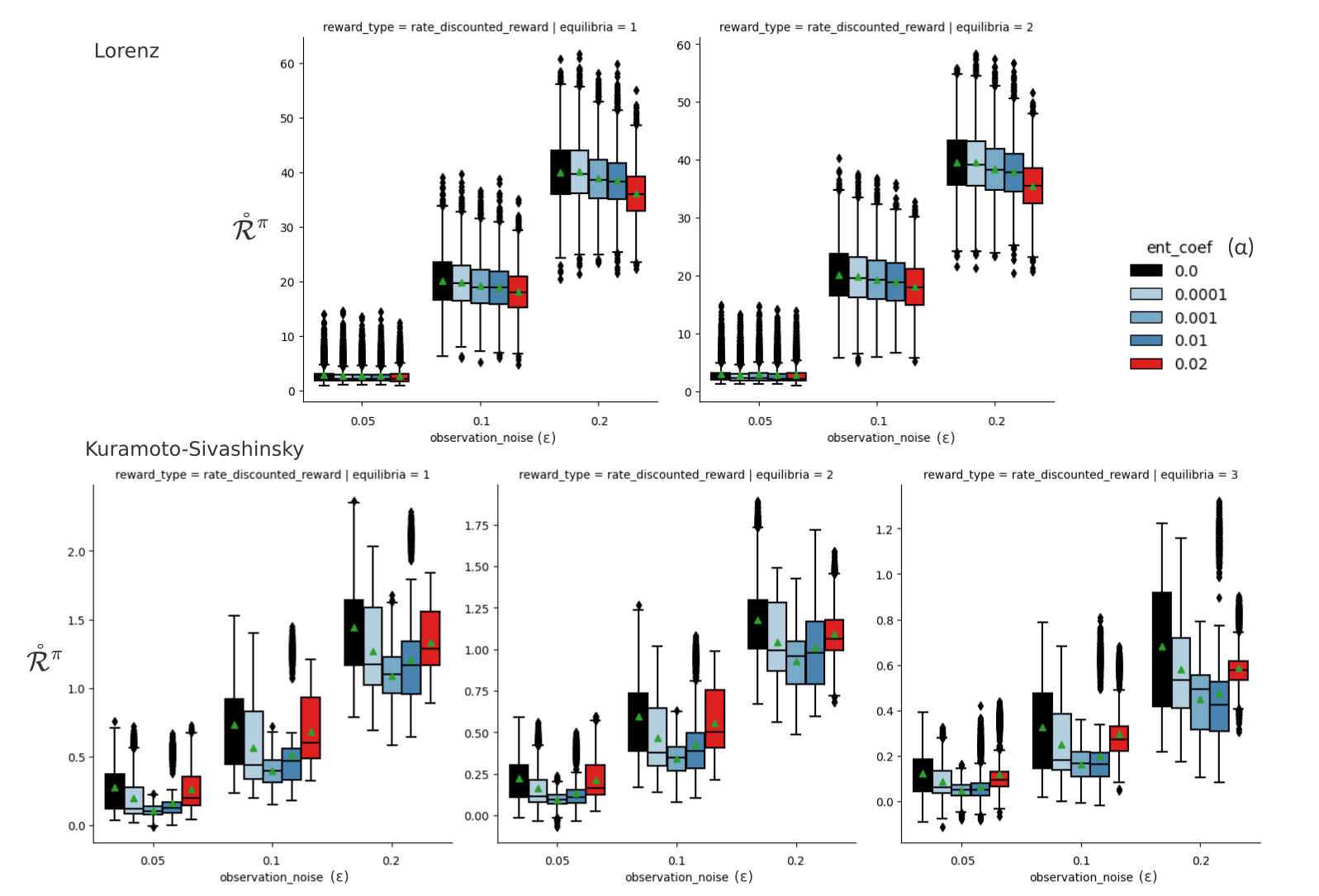}
        \caption{Distributional representation
        of the rate of excess risk under noise $\mathring{\mathcal{R}}^{\pi}$
            conditioned on the $\alpha^i$ used during optimisation
            for different initial state distribution
            $X_0 \sim \mathcal{N}(x_e^{*},\, \sigma_{e}^2 I_d)$.
                {\color{black}
            Each of the rows corresponds to one of the dynamical systems
            of interest.
            Each of the columns corresponds to one of
            the initial state distributions of interest.
            There are two non-zero fixed points (equilibria) $x_e^*$ for Lorenz and three for KS.
            From top to bottom: KS; Lorenz.\\
            For each box plot,
            three intensities $\sigma_Y$ for the observation noise $\epsilon$
            are evaluated.
            As expected,
            when the uncertainty regarding the observable $Y$ increases
            through the variance $\sigma_Y$
            of the observation signal noise $\epsilon$,
            the policy performance decreases globally
            ($\mathring{\mathcal{R}}^{\pi}$ increases).
            Moreover, the rate of excess risk under noise tends to decrease when $\alpha^i$ increases in the Lorenz case, whereas it decreases up to a certain entropy threshold for KS before increasing again.}}
        \label{fig:all_dynamics_boxplot_robustness_rate}
    \end{figure}

    First and foremost, the results shown in
    Figure~\ref{fig:all_dynamics_boxplot_robustness_rate} indicate
    that the noise introduction to the system observable $Y$ of KS and Lorenz
    leads to a global decrease in performance, as expected.

    The robustness to noise contamination of the two systems is improved by initialising the policy optimisation procedure up to a certain intermediate threshold of the entropy coefficient $\alpha^i > 0$.
    Once this value is reached, two respective behaviours are observed depending on the system.
    In the case of the Lorenz dynamics, the robustness continues to improve after this entropy threshold, whereas the opposite trend is observed for KS (particularly with the maximal entropy coefficient chosen).

    Hence, the sole introduction of entropy-regularisation
    in the objective function impacts the robustness.
    This behaviour difference between Lorenz and KS might be explained by the variability of the optimisation landscapes that can be observed with respect to the chosen underlying dynamics as underlined in~\cite{ahmed2019}.

    \subsection[Maximum entropy as a norm-based regularisation on the policy]
    {Maximum entropy as a norm-based regularisation on the policy}
    \label{subsec:maximum-entropy-as-a-norm-based-regularisation-on-the-policy}
    Norm-based complexity measures introduced
    in Section~\ref{subsec:complexity-measures-for-po-mdp-with-gaussian-noise}
    are now evaluated.
    For a complexity measure $\mathcal{M}$ to be considered significant,
    it should be correlated with the robustness of the model.

\begin{figure}[t]
        \centering
        \includegraphics[scale=0.23]{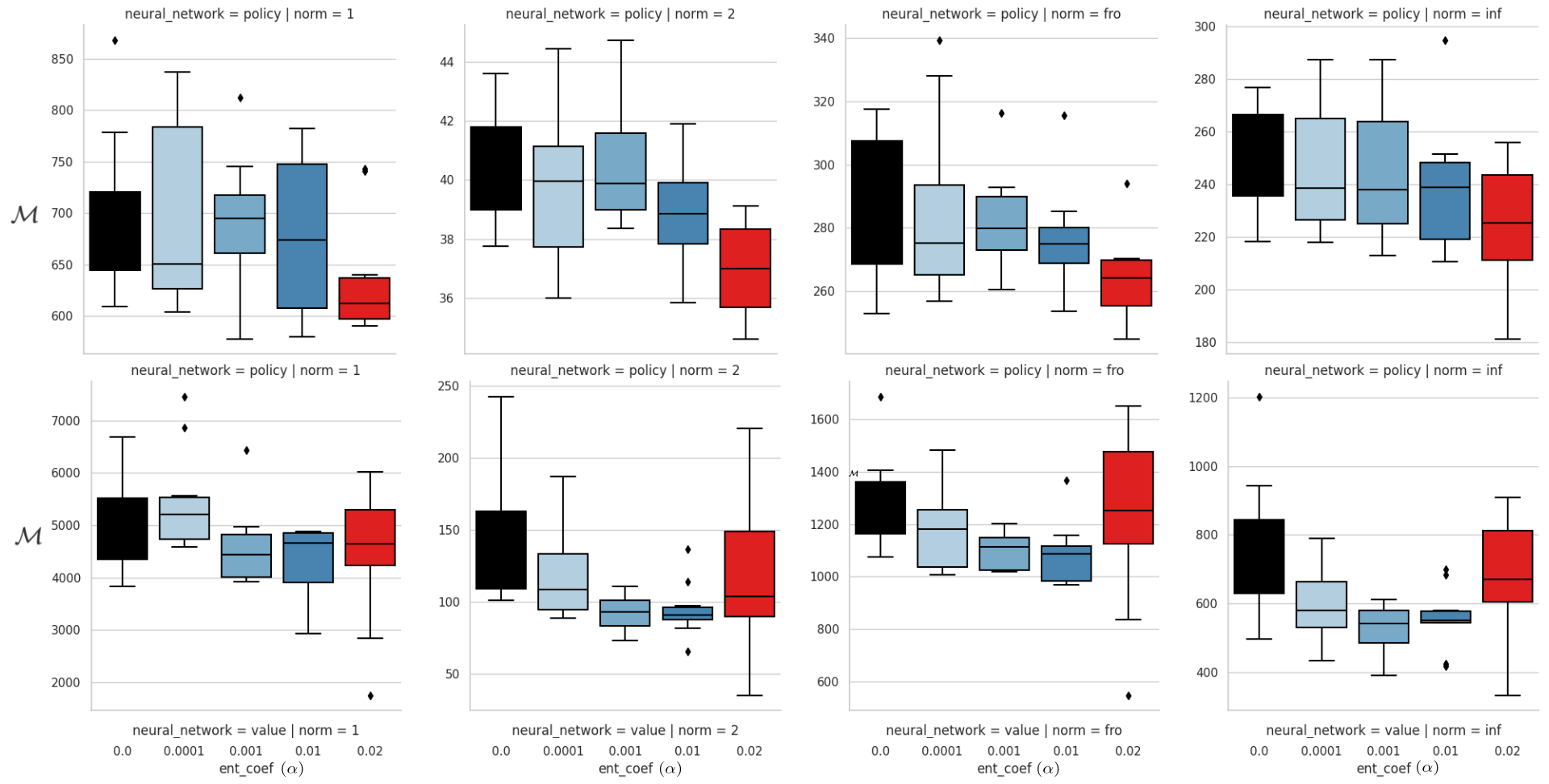}
        \caption{Measures of complexity $\mathcal{M}(\pi_\theta, \mathcal{D}) = \Pi^l_{i=1} \| \theta_\mu^i \|_p$ with $p = 1,\, 2,\, \infty,\, F$ conditioned on the $\alpha^i$ used during optimisation.
                {\color{black} Each row
            corresponds to one of the dynamical systems of interest
            while column represents a different norm order $p$.
            From top to bottom: Lorenz and KS.\\
            For the Lorenz case, the barycenters of the measures tend to decrease when $\alpha^i$ increases.
            Regarding KS, passing a threshold, the complexity increases again with the entropy.
            In addition,
            the measures are much more concentrated when $\alpha^i > 0$.
            For $p = 2,\, F$, the separation of the measures w.r.t.\
            the different $\alpha^i$ is more pronounced.}}

        \label{fig:all_dynamics_lipschitz_product}
    \end{figure}

    Accordingly,
    the different norm-based measures presented
    in Section~\ref{subsec:complexity-measures-for-po-mdp-with-gaussian-noise}
    are estimated.
    Figure~\ref{fig:all_dynamics_lipschitz_product} shows
    the layer-wise product norm of the policy actor network parameters
    ($\mathcal{M}(\pi_\theta, \mathcal{D}) = \Pi^l_{i=1} \| \theta_\mu^i \|_p$)
    w.r.t.\ to their associated entropy coefficient $\alpha^i$
    for all the 50 independently trained models.

    Again, policies obtained with initial $\alpha^i > 0$ exhibit a trend toward decreasing complexity measure values as $\alpha$ increases up to a certain threshold of the entropy coefficient.
    Similarly to Section~\ref{subsec:entropy-regularisation-induces-noise-robustness}, the complexity measure continues to decrease after surpassing this threshold for the Lorenz system.
    On the other hand, in the KS case, $\mathcal{M}(\pi_\theta, \mathcal{D})$ increases again once its entropy threshold is reached, notably for the larger entropy coefficient.

    Moreover, the measures tend to be much more concentrated when $\alpha^i > 0$, especially in the case of KS (except for the higher $\alpha^i$).

    This may indicate that the entropy regularisation acts on the uncertainty
    of the policy parameters.
    Likewise, similar observations can be made for the total norm
    of the parameters but are not introduced here for the sake of brevity.

    Consequently, this experiment highlights an existing correlation
    between maximum entropy regularisation and norm-based complexity measures.
    As this complexity measure
    is linked to the Lipschitz continuity of the policy,
    one might wonder if the regularity of the policy is more directly impacted.
    This is the purpose of the next subsection.

    \subsection[Maximum entropy reduces the average Fisher-Information]
    {Maximum entropy reduces the average Fisher-Information}
    \label{subsec:maximum-entropy-reduces-the-average-fisher-information}
    Another regularity measure is considered:
    the average trace of the Fisher information
    ($\mathcal{M}(\pi_\theta, \mathcal{D})
    = Tr( \mathcal{I}\left(\theta_\mu \right))
    = Tr(-\ \mathbb{E}^{X \sim \rho, U \sim \pi_\theta(\cdot | X)}
    \left[ \nabla^2_{\theta_\mu} \log \pi_\theta(U \mid X) \right])$).
    As discussed in~\ref{subsubsec:flatness-based-complexity-measures},
    this quantity reflects the regularity of the policy and might be related
    to the flatness of the local minima of the objective function.

\begin{figure}[t]
        \centering
        \includegraphics[scale=0.28]{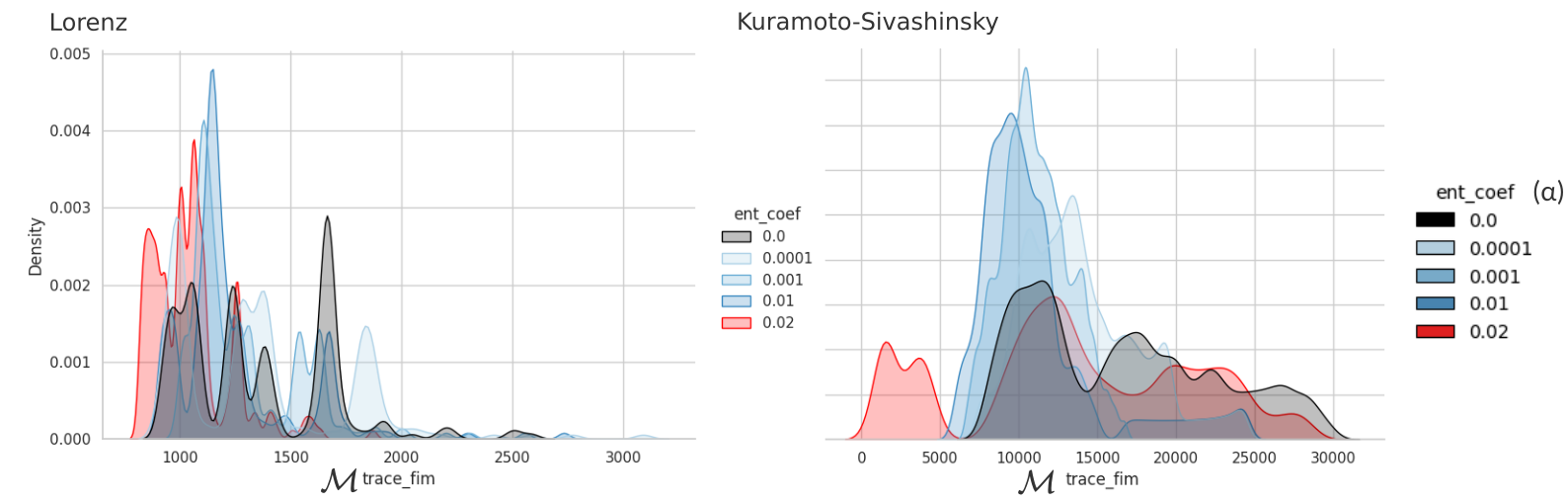}
        \caption{Distribution of the trace of the (conditional) Fisher information of the numerical optimal solution $\theta^*_{\mu, \alpha^i}$ for the policy w.r.t.\ the $\alpha^i$ used during optimisation.
                {\color{black}
            From left to right: Lorenz and KS environments.
            Colours: control experiment $\alpha^i = 0$ (black);
            intermediate entropy level $\alpha^i$ (blue);
            largest $\alpha^i$ (red). \\
            A skewed distribution towards (relatively) larger values
            is observed for all controlled dynamical systems. 
            Moreover,
            those right tails exhibit high kurtosis, especially
            for the control experiment (black)
            and the model with the larger entropy coefficient (red)
            for the KS system.
            Finally, solutions with intermediate entropy levels (blue)
            are much more concentrated - have lower variance than the others.
            About Lorenz, the barycenter of the more robust model (red) is shifted
            towards lower values than the others.
            }}
        \label{fig:all_dynamics_trace_fisher_information}
    \end{figure}

    Figure~\ref{fig:all_dynamics_trace_fisher_information}
    shows the distribution under $\pi_\theta$ of the trace
    of the state conditional Fisher Information
    of the numerical optimal solution $\theta^*_{\mu, \alpha^i}$
    for the policy w.r.t.\ the $\alpha^i$ used during optimisation.
    In other words, a kernel density estimator of the distribution of
    $Tr( \mathcal{I}(\pi_{\theta^*_{\mu, \alpha^i}}(~\cdot \mid X) ))$
    when $X \sim \rho^{\pi_{\theta^*}}$ is represented.
    The results of this experiment suggest first, this distribution
    is skewed negatively and has a fat right tail.
    This means some regions of the support of $\rho^{\pi_{\theta^*}}$
    provide FIM trace with extreme positive values,
    meaning the regularity of the policy may be poor
    in these regions of the state space. \\
    A comparison of the distribution w.r.t.\ the different $\alpha^i$
    sheds further light on the relation between robustness and regularity.
    In fact, there appears to be a correspondence between the robustness,
    as indicated by the rate of excess risk under noise
    $\mathring{\mathcal{R}}^{\pi}$
    shown in Figure~\ref{fig:all_dynamics_boxplot_robustness_rate}
    and the concentration of the trace distribution toward larger values
    (\textit{i.e.} more irregular policies) when the model is less robust.

    Meanwhile,
    under the considerations
    of~\ref{subsubsec:flatness-based-complexity-measures}
    and since it is known that entropy regularisation favours flat minima
    in RL~\cite{ahmed2019},
    these experimental results support the hypothesis
    of an existing relationship between robustness,
    objective function flatness around the solution $\theta^*$
    and conditional Fisher information of $\theta^*$.

    For a complementary point of view, a supplementary experiment regarding the sensitivity of the policy updates during training w.r.t.\ to different level of entropy is also presented in Appendix~\ref{sec:appendix-weights-sensitivity-during-training}.

    \section[discussion]{Discussion}\label{sec:discussion}

    In this paper, the question of the robustness of maximum entropy policies
    under noise is studied.
    After introducing the notion of complexity measures from the statistical
    learning theory literature, numerical evidence supports
    the hypothesis that maximum entropy regularisation
    induces robustness under noise.
    Moreover,
    norm-based complexity measures are shown to be correlated
    with the robustness of the model.
    Then,
    the average trace of the Fisher Information is shown to be a relevant
    indicator of the regularity of the policy.
    This suggests the existence of a link between robustness,
    regularity and entropy regularisation.
    Finally, this work contributes to bringing statistical learning concepts
    such as flatness into the field of Reinforcement Learning.
        {\color{black}
    New algorithms or metrics, such as in the work of~\cite{lecarpentier2021},
    may be built upon notions of regularity, \textit{e.g.},
    Lipschitz continuity, flatness or Fisher Information of the parameter
    in order to achieve robustness.}

    \begin{credits}
        \subsubsection{\ackname}
        The authors acknowledge the support
        of the French Agence Nationale de la Recherche (ANR),
        under grant ANR-REASON (ANR-21-CE46-0008)\@.
        This work was performed using HPC resources
        from GENCI-IDRIS (Grant 2023-[AD011014278]).

        \subsubsection{\discintname}
        The authors have no competing interests to declare that are relevant to the content of this article.
    \end{credits}

    \appendix

    \section{Weights sensitivity during training}\label{sec:appendix-weights-sensitivity-during-training}

    This section is intended to provide complementary insights on the optimisation landscape induced by the entropy coefficient $\alpha$ during training from the \emph{conservative} or \emph{trust region} policy iteration point of view~\cite{kakade2002,schulman2015}.

    Let $\left( \theta^\alpha_m \right)_{m=1}^{m_\mathcal{D}}$ be the sequence of weights of the policy during the training of the model for some initial entropy coefficient $\alpha$.
    The conditional Kullback-Leibler divergence between the policy identified by the parameters $\theta^\alpha_m$ and the subsequent policy defined by the parameters $\theta^\alpha_{m + 1}$ is given by \\
    $\overline{D}_{KL}\left( \theta^\alpha_m,\, \theta^\alpha_{m + 1} \right)
    = \mathbb{E}^{X \sim \rho}
    \left[
        \int_{\mathcal{U}}
        \log\left( \frac{\pi_{\theta^\alpha_m}\left(du \mid X\right)}{\pi_{\theta^\alpha_{m + 1}}\left(du \mid X\right)}\right)
        \pi_{\theta^\alpha_{m + 1}}(du \mid X)
        \right]$. \\
    The above quantity is a measure of the divergence from the policy at time $m$ to the policy at time $m + 1$.
    Thus it may provide information on the local stiffness of the optimisation landscape during training.

\begin{figure}[h]
        \centering
        \begin{subfigure}{.5\textwidth}
            \centering
            \includegraphics[width=.8\linewidth]{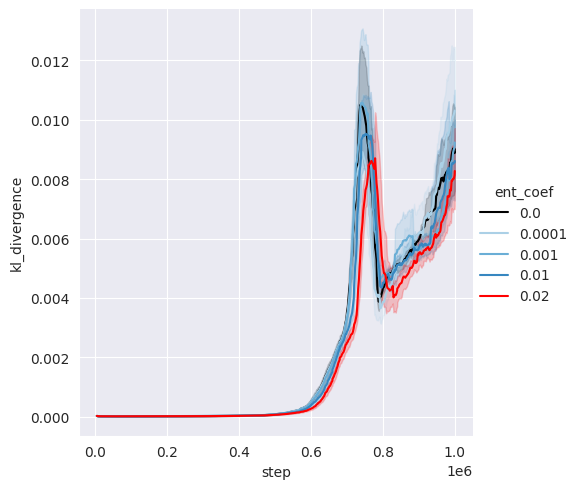}
            \caption{Lorenz}
            \label{fig:kl_divergence_lorenz}
        \end{subfigure}%
        \begin{subfigure}[Kuramoto-Sivashinsky]{.5\textwidth}
            \centering
            \includegraphics[width=.8\linewidth]{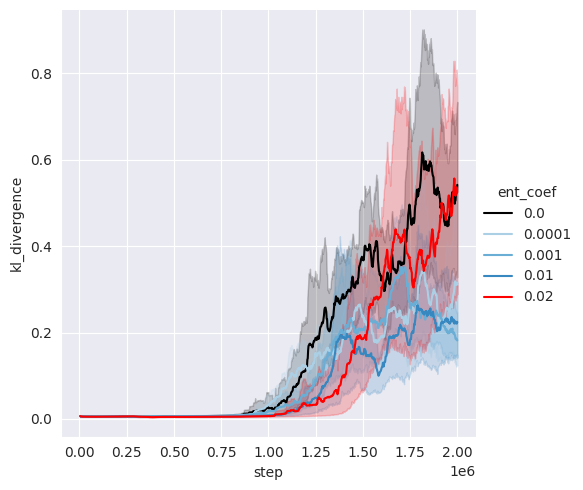}
            \caption{Kuramoto-Sivashinsky}
            \label{fig:kl_divergence_ks}
        \end{subfigure}
        \caption{Evolution of $\overline{D}_{KL}\left( \theta^\alpha_m,\, \theta^\alpha_{m + 1} \right)$ during training for the Lorenz and KS controlled differential equations.
        For Lorenz, the maximal divergence is reached for the optimisation performed with $\alpha^i = 0$ and the second lowest $\alpha^i$.
        Regarding KS, the highest divergence values are observed for $\alpha^i = 0$ and the maximal entropy coefficient.}
        \label{fig:kl_divergence_evolution}
    \end{figure}

    Figure~\ref{fig:kl_divergence_evolution} shows the evolution of the Kullback-Leibler divergence between two subsequent policies during training for the Lorenz and KS controlled differential equations.
    Regarding the Lorenz system, the maximal divergence is reached for the optimisation performed with the two lowest $\alpha^i$ while increasing entropy seems to slightly reduce the divergence.
    On the other hand, the highest divergence values observed for the KS system are reached for $\alpha^i = 0$ and the maximal entropy coefficient.
    This observation is coherent with the results of the previous sections and suggests that the entropy coefficient $\alpha$ impacts the optimisation landscape during training.

    Interesting questions regarding the optimisation landscape and its link with the Fisher Information (through the point of view of Information Geometry~\cite{amari1998}) are raised by the results of this section but are left for future work.


    \bibliographystyle{splncs04}
    \bibliography{entropy_regularity_report_springer_final}
%
%
%
%
%
\end{document}